\pdfoutput=1
\documentclass[11pt]{article}

\usepackage{ACL2023}

\usepackage{times}
\usepackage{latexsym}

\usepackage[T1]{fontenc}
\usepackage[utf8]{inputenc}

\usepackage{microtype}

\usepackage{inconsolata}

\usepackage{graphicx}
\usepackage{amsmath}
\usepackage{amssymb}

\title{Semantic Structure Enhanced Event Causality Identification}

\author{Zhilei Hu,  Zixuan Li$^{*}$, Xiaolong Jin$^{*}$, Long Bai, Saiping Guan, \\
{\bf Jiafeng Guo} \and {\bf Xueqi Cheng} \\
        School of Computer Science and Technology, University of Chinese Academy of Sciences; \\ 
        CAS Key Laboratory of Network Data Science and Technology, \\
        Institute of Computing Technology, Chinese Academy of Sciences. \\
        \texttt{\{huzhilei19b, lizixuan, jinxiaolong, bailong18b, guansaiping\}@ict.ac.cn} \\
        \texttt{\{guojiafeng, cxq\}@ict.ac.cn}
        }

\newcommand \footnoteONLYtext[1]
{
	\let \mybackup \thefootnote
	\let \thefootnote \relax
	\footnotetext{#1}
	\let \thefootnote \mybackup
	\let \mybackup \imareallyundefinedcommand
}

\begin{document}
\graphicspath{{./}}
\maketitle
\footnoteONLYtext{$^{*}$Corresponding authors.}

\begin{abstract}

Event Causality Identification (ECI) aims to identify causal relations between events in unstructured texts.
This is a very challenging task, because causal relations are usually expressed by implicit associations between events.
Existing methods usually capture such associations by directly modeling the texts with pre-trained language models, which underestimate two kinds of semantic structures vital to the ECI task, namely, event-centric structure and event-associated structure.
The former includes important semantic elements related to the events to describe them more precisely, while the latter contains semantic paths between two events to provide possible supports for ECI. 
In this paper, we study the implicit associations between events by modeling the above explicit semantic structures, and propose a \textbf{Sem}antic \textbf{S}tructure \textbf{In}tegration model (\textbf{SemSIn}).
It utilizes a GNN-based event aggregator to integrate the event-centric structure information, and employs an LSTM-based path aggregator to capture the event-associated structure information between two events.
Experimental results on three widely used datasets show that SemSIn achieves significant improvements over baseline methods.

\end{abstract}

\section{Introduction}

\begin{figure}[t]
  \centering
  \includegraphics[width=\linewidth]{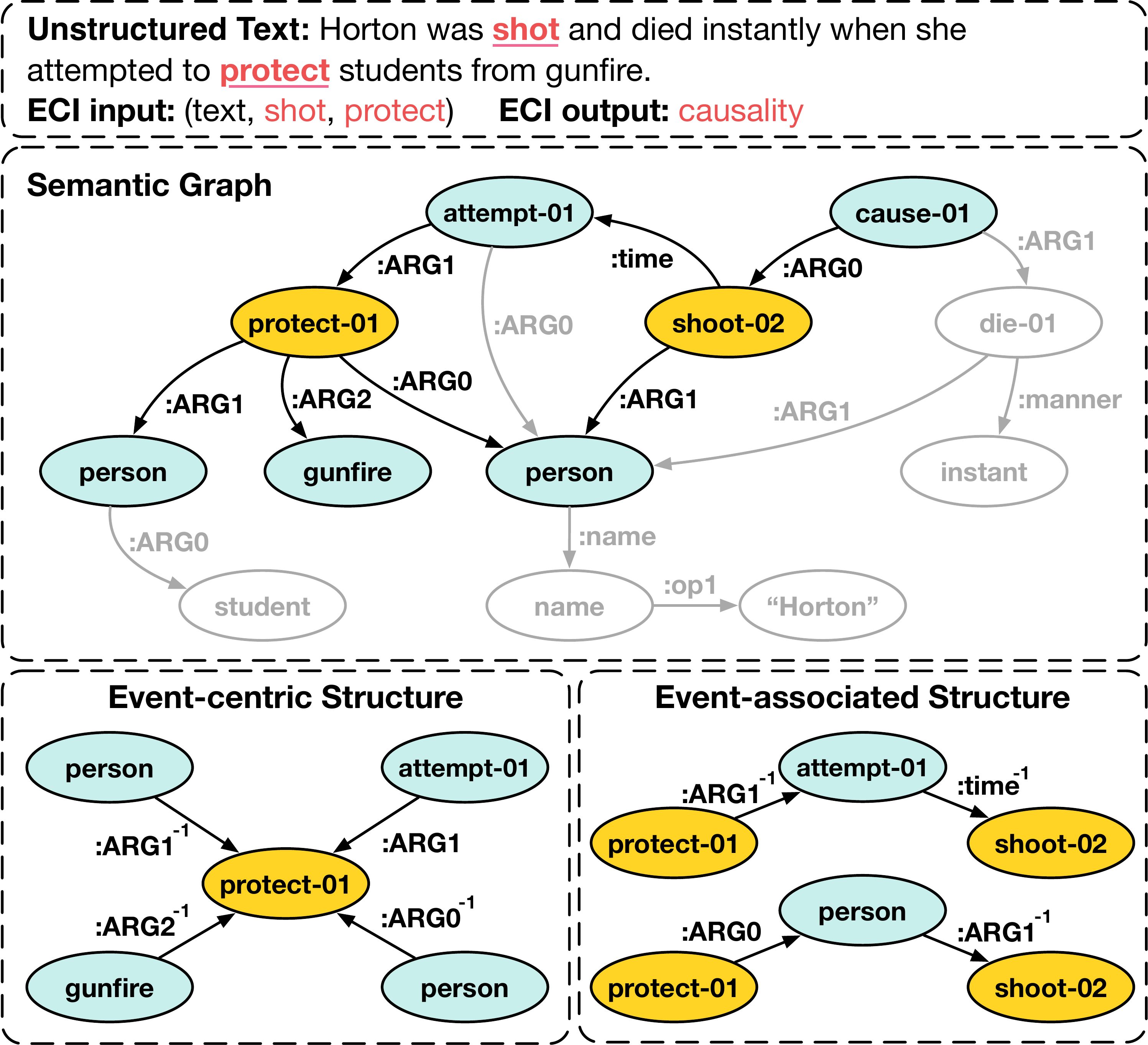}
  \caption{An example of the ECI task, as well as the semantic graph and semantic structures corresponding to the unstructured text. 
  The orange nodes denote events.}
  \label{fig:example}
\end{figure}

Event Causality Identification (ECI) is an important task in natural language processing that seeks to predict causal relations between events in texts.
As shown in the top of Figure~\ref{fig:example}, given the unstructured text and event pair (\emph{\textbf{shot}, \textbf{protect}}), an ECI model needs to identify that there exists a causal relation between two events, i.e., $\emph{\textbf{protect}} \stackrel{cause}{\longrightarrow} \emph{\textbf{shot}}$.
ECI is an important way to construct wide causal connections among events, which supports a variety of practical applications, such as event prediction~\citep{Hashimoto2019WeaklySupervisedMultilingual}, reading comprehension~\citep{Berant2014ModelingBiologicalProcesses}, and question answering~\citep{Oh2013WhyQuestionAnsweringUsing, Oh2017MultiColumnConvolutionalNeural}.

ECI is challenging because most causal relations are expressed by texts implicitly, which requires the model to understand the associations between two events adequately.
Existing methods directly model the texts with the Pre-trained Language Model (PLM)~\citep{Liu2020KnowledgeEnhancedEvent, Cao2021KnowledgeEnrichedEventCausality, Zuo2020KnowDisKnowledgeEnhanced, Zuo2021LearnDALearnableKnowledgeGuided}.
They mainly rely on the ability of PLMs, which cannot capture associations between events comprehensively.
To enrich the associations between events, some methods~\citep{Liu2020KnowledgeEnhancedEvent, Cao2021KnowledgeEnrichedEventCausality} introduce external knowledge, such as events in ConceptNet~\citep{Speer2017ConceptnetOpenMultilingual} that are related to focused events.
Other methods~\citep{Zuo2020KnowDisKnowledgeEnhanced, Zuo2021LearnDALearnableKnowledgeGuided} utilize the data augment framework to generate more training data for the model.
However, the above methods are far from fully modeling the associations among events in the texts.

Actually, texts contain rich semantic elements and their associations, which form graph-like semantic structures, i.e., semantic graphs.
Figure~\ref{fig:example} shows the semantic graph generated by the Abstract Meaning Representation (AMR)~\citep{Banarescu2013AbstractMeaningRepresentation} parser for the corresponding text, where the nodes indicate events, entities, concepts, and other semantic elements, while edges with semantic roles describe the associations among semantic elements.
For example, ``protect-01'' indicates the specific sense of the verb ``protect'' in the PropBank~\citep{Palmer2005PropositionBankAnnotated}~\footnote{A corpus annotated with verbs and their semantic roles.}.
“ARG0”, “ARG1” and “ARG2” indicate different semantic roles.
In this semantic graph, we exploit two kinds of structures vital to the ECI task, namely, event-centric structure and event-associated structure.
As shown in the bottom left of Figure~\ref{fig:example}, the event-centric structure consists of events and their neighbors, which describes events more precisely by considering their arguments and corresponding roles.
For example, besides event ``protect-01’’, ``person (Horton)'' and ``person (student)'' are also important semantic elements, and their corresponding semantic roles can supply the extra information for the event.
As shown in the bottom right of Figure~\ref{fig:example}, the event-associated structure contains semantic paths between two events, and each path contains the core semantic elements.
The composition of these elements indicates the possible semantic relations between events and provides supports for ECI. 
For example, the path ``$\text{protect-01} \xrightarrow{\text{:ARG0}} \text{person} \xrightarrow{\text{:ARG1}^{-1}} \text{shoot-02}$'' indicates that ``person (Horton)'' protects somebody first and then was shot. 
Events ``protect-01'' and ``shoot-02'' share the same participant, and there may exist a causal relation between them.

To make use of the above semantic structures in texts to carry out the ECI task, we propose a new \textbf{Sem}antic \textbf{S}tructure \textbf{In}tegration model (SemSIn).
It first employs an AMR parser to convert each unstructured text into a semantic graph and obtains the above two kinds of semantic structures in that graph.
For the event-centric structure, SemSIn adopts an event aggregator based on Graph Neural Networks (GNN). 
It aggregates the information of the neighboring nodes to the event nodes to obtain more precise representations of the events.
For the event-associated structure, SemSIn utilizes a path aggregator based on Long Short-Term Memory (LSTM) network. 
It encodes the compositional semantic information in the paths and then integrates the information of multiple paths with an attention mechanism.
With the above representations of the events and paths as input, SemSIn conducts ECI with a Multi-Layer Perceptron (MLP).

In general, the main contributions of this paper can be summarized as follows:
\begin{itemize}
\item 
We exploit two kinds of critical semantic structures for the ECI task, namely, event-centric structure and event-associated structure. They can explicitly consider the associations between events and their arguments, as well as the associations between event pairs.
\item We propose a novel Semantic Structure Integration (SemSIn) model, which utilizes an event aggregator and a path aggregator to integrate the above two kinds of semantic structure information.
\item According to experimental results on three widely used datasets, SemSIn achieves 3.5\% improvements of the F1 score compared to the state-of-the-art baselines. 
\end{itemize}

\section{Related Work}

\begin{figure*}[ht]
  \centering
  \includegraphics[width=\textwidth]{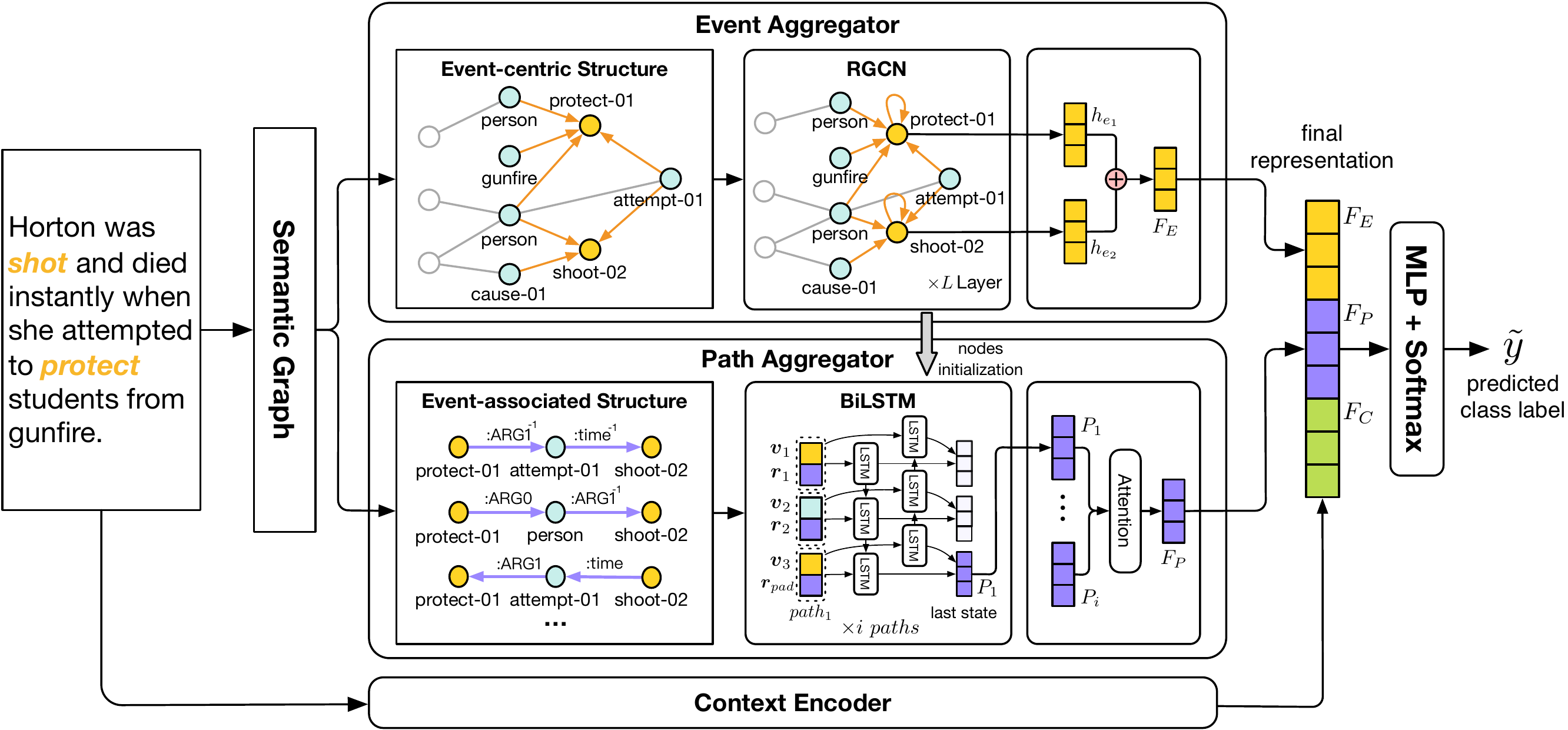}
  \caption{An illustration diagram of the proposed SemSIn model.}
  \label{fig:model}
\end{figure*}

Identifying causal relations between events has attracted extensive attention in the past few years.
Early methods mainly rely on the causal association rule~\citep{Beamer2009UsingBigramEvent, Do2011MinimallySupervisedEvent} and causal patterns~\citep{Hashimoto2014FutureScenarioGeneration, Riaz2010AnotherLookCausality, Riaz2014IndepthExploitationNoun, Hidey2016IdentifyingCausalRelations}.
Some following methods exploit lexical and syntactic features to improve performance~\citep{Riaz2013BetterUnderstandingCausality, Riaz2014RecognizingCausalityVerbnoun}.

Recently, most of works apply PLM to conduct ECI~\citep{Liu2020KnowledgeEnhancedEvent, Cao2021KnowledgeEnrichedEventCausality, Zuo2021LearnDALearnableKnowledgeGuided}.
Although PLM has a strong ability for capturing the associations among tokens in the texts, they are not capable of this task because the associations between events are implicit.
To enhance PLM, recent works try to introduce external knowledge.
\citet{Liu2020KnowledgeEnhancedEvent} proposed a method to enrich the representations of events using commonsense knowledge related to events from the knowledge graph ConceptNet~\citep{Speer2017ConceptnetOpenMultilingual}.
\citet{Cao2021KnowledgeEnrichedEventCausality} further proposed a model to exploit knowledge connecting events in the ConceptNet for reasoning.
\citet{Zuo2021LearnDALearnableKnowledgeGuided} proposed a data augmented method to generate more training samples.
Instead of introducing external knowledge to enhance the abilities of the ECI model, we attempt to dive deep into the useful semantic structure information in the texts.

\section{The SemSIn Model}

In this section, we introduce the proposed SemSIn model.
Figure~\ref{fig:model} illustrates the overall architecture of the SemSIn model.
Given an input text, SemSIn first uses a pre-trained AMR parser to obtain the corresponding semantic graph of the text. 
Then, the event-centric structure and the event-associated structure constructed from the semantic graph, as well as the original text, are fed into the following three components respectively: 
(1) Event aggregator aggregates the event-centric structure information into the representation of the event pair.
(2) Path aggregator captures the event-associated structure information between two events into the path representation. 
(3) Context encoder encodes the text and obtains the contextual representation of the event pair.
With the above representations as input, SemSin conducts binary classification to get the final results with an MLP layer.
Next, we will first introduce the construction process of the semantic graph and then present these three main components in detail.

\subsection{Semantic Graph Construction}
The core motivation of SemSIn is to model the implicit associations between events by introducing explicit semantic structures. 
To get explicit semantic structures from texts, SemSIn employs an AMR parser to convert the original text into an AMR graph, which contains fine-grained node and edge types~\citep{Zhang2021AbstractMeaningRepresentation}.

In the AMR graph, the nodes indicate specific semantic elements and the edges indicate the semantic roles among them. 
Table~\ref{table:amr-roles} lists the used semantic roles in AMR graph.
We then add inverse edges to all the edges in the AMR graph to form the final semantic graph, making it reachable between any two nodes. Formally, a semantic graph is defined as $G = (V, E, R)$, where $V$, $E$ and $R$ are the sets of nodes, edges and role types, respectively.

\begin{table}[]
    \centering

    \begin{tabular}{c | c }
    \hline
    \textbf{Semantic Roles} & \textbf{Types} \\
    \hline
    ARG0, ARG1, ARG2, $\cdots$ & Core Roles \\
    op1, op2, op3, op4 & Operators \\
    manner, instrument, topic, $\cdots$ & Means \\
    time, year, weekday, $\cdots$ & Temporal \\
    Other semantic roles & Others \\
    \hline
    \end{tabular}
    \caption{Semantic roles in the AMR graphs~\citep{Zhang2021AbstractMeaningRepresentation}}
    \label{table:amr-roles}
\end{table}

\subsection{Event Aggregator}

Identifying the causal relation between two events requires the model to comprehensively understand what each event describes. 
Existing methods use the event mentions in the text to represent the events, which cannot highlight the semantic elements related to the events.
Besides the event mentions, events usually have associations with their arguments mentioned in the text. 
Similarly, event arguments also have associations with some related semantic elements.
Therefore, to model this kind of association, SemSIn obtains the event-centric structure from the constructed semantic graph by simply using the $L$-hop subgraph of the focused event, where $L$ is a hyperparameter.

\textbf{Node Representation Initialization:} To initialize the representations of nodes in the event-centric structure, a rule-based alignment tool~\footnote{RBW Aligner in \url{https://github.com/bjascob/amrlib}} is first employed to align AMR nodes to the tokens in the text. 
For the AMR nodes that have the corresponding tokens in the text, their initialized representations are obtained by averaging the representation vectors of all tokens aligned to the nodes.
For example, given a node, the start and end positions of its corresponding tokens are $a$ and $b$, respectively. 
Its representation vector is calculated by:
\begin{equation}
   \boldsymbol{h} = \frac{1}{|b-a+1|} \sum_{k=a}^{b} \boldsymbol{x}_k,
\end{equation}
where $\boldsymbol{x}_k$ is the representation of the token $k$.
A PLM, BERT~\citep{Devlin2019BERTPretrainingDeep}, is applied to encode the sequence of tokens.
For those nodes without corresponding tokens in the original text (i.e., auxiliary nodes added by the AMR parser, such as ``name'' and ``cause-01'' in Figure~\ref{fig:example}), their representations are randomly initialized.

\textbf{Semantic Information Aggregation}: The graph convolutional network has the property of aggregating the information of neighbor nodes to the specific node, which is suitable to model the event-centric structure.
In addition, the types of edges in the semantic graph also contain special information that can be used to distinguish the relations between nodes.
Therefore, we apply a Relational Graph Convolutional Network (RGCN)~\citep{Schlichtkrull2018ModelingRelationalData} to aggregate semantic information from $L$-hop neighbors of the focused events. Specifically, the message passing at layer $l\in [0, L-1]$ is conducted as follows:
\begin{equation}
    \boldsymbol{h}_{i}^{l+1} = \sigma \left( \sum_{r\in R} \sum_{j\in N_i^r} \frac{1}{c_{i,r}} \boldsymbol{W}_{r}^{l} \boldsymbol{h}_{j}^{l} + \boldsymbol{W}_{0}^{l} \boldsymbol{h}_{i}^{l} \right),
\end{equation}
where $R$ denotes the set of the role types; $N_i^r $ denotes the set of the neighbors of node $i$ under relation $r \in R$;
$c_{i,r}$ is a normalization constant equal to $|N_i^r|$;
$\boldsymbol{h}_{i}^{l}$ and $\boldsymbol{h}_{j}^{l}$ denote the $l^{th}$ layer representations of the nodes $i$ and $j$, respectively;
$\boldsymbol{W}_{r}^{l}$ and $\boldsymbol{W}_{0}^{l}$ are the weight matrices for aggregating features from different relations and self-loop in the $l^{th}$ layer;
$\sigma$ is an activation function (e.g., ReLU); $\boldsymbol{h}_{i}^{0}$ and $\boldsymbol{h}_{j}^{0}$ are the initialized representations of the nodes introduced above.
After aggregating the event-centric structure information, the representations of $e_1$ and $e_2$ are denoted as $\boldsymbol{h}_{e_1}$ and $\boldsymbol{h}_{e_2}$. 
In addition, to eliminate the effect of the relative position of the two events, we sum up the representations of the two events to obtain $\boldsymbol{F}_{E}^{(e_1, e_2)}$, the representation of the event pair,
\begin{equation}
    \boldsymbol{F}_{E}^{(e_1, e_2)} = \boldsymbol{h}_{e_1} + \boldsymbol{h}_{e_2}.
\end{equation}

\subsection{Path Aggregator}

Besides the associations between events and their arguments, identifying the causal relation requires the model to discover the association between two events. The paths in the semantic graph between two events can reflect this kind of association. SemSIn thus first finds paths between two events in the semantic graph to form the event-associated structure. Then, SemSIn encodes it via BILSTM and path attention to get the representations of the paths. 

 With the intuition that the fewer hops in the path, the stronger information it contains to reflect the association between two events, we choose the shortest path between two event nodes in the semantic graph to form the event-associated structure. 
 This operation can avoid introducing redundant information and improve efficiency. 
Besides, we add the reverse path for each semantic path.
Formally, if there is a path denoted as $(v_1, r_1, v_2, \cdots, v_{n-1}, r_{n-1}, v_n)$, the corresponding reverse path is $(v_n, r_{n-1}, v_{n-1}, \cdots, v_2, r_1, v_1)$.

\textbf{Path Encoding:}
The compositional semantic information of the semantic elements and roles in paths can provide possible supports to the causal relation. Recently, recurrent neural networks have been widely used in processing sequence data such as path information~\citep{Wang2019ExplainableReasoningKnowledge, Huang2021PathenhancedExplainableRecommendation}.
Therefore, we apply a BiLSTM to better encode each path in the event-associated structure and output its representations.
Here, the initialized representations of all nodes are obtained by applying the RGCN to the whole semantic graph, while the representations of relations are randomly initialized and updated during the training process.
To convert multi-hop paths into a sequence of vectors, we concatenate node and relation representation vectors as the input at each state.
For example, the sequence is organized as $\left[ (\boldsymbol{v}_1, \boldsymbol{r}_1); (\boldsymbol{v}_2, \boldsymbol{r}_2); \cdots; (\boldsymbol{v}_n, \boldsymbol{r}_{pad}) \right]$, 
where $\boldsymbol{v}_i$ denotes the representation of the node $i$; $\boldsymbol{r}_i$ denotes the representation of the relation $i$; $\boldsymbol{r}_{pad}$ denotes the representation of the special PAD relation added to the last state.
Then, the representation $\boldsymbol{P}_i$ of this path can be obtained:
\begin{equation}
    \boldsymbol{P}_i = \mathrm{BiLSTM} \left[(\boldsymbol{v}_1, \boldsymbol{r}_1); \cdots; (\boldsymbol{v}_n, \boldsymbol{r}_{pad}) \right].
\end{equation}

\textbf{Path Attention:}
There may exist multiple paths with the same shortest length. 
Different paths reflect different semantic information.
Thus, to distinguish the importances of different paths, SemSIn adopts an attention mechanism to integrate the information of multiple paths.
The query for attention is the representation of the event pair $\boldsymbol{F}_E^{(e_1, e_2)}$, which is obtained from the event aggregator. 
Both key and value are the representation $\boldsymbol{P}_i$ of the path:
\begin{equation}
    \alpha_{i} = \frac{(\boldsymbol{F}_E^{(e_1, e_2)} \boldsymbol{W}_Q)(\boldsymbol{P}_{i} \boldsymbol{W}_K)^T}{\sqrt{d_k}},
\end{equation}
\begin{equation}
    \boldsymbol{F}_P^{(e_1, e_2)} = \sum_{i} \mathrm{Softmax}(\alpha_{i}) (\boldsymbol{P}_{i} \boldsymbol{W}_V),
\end{equation}
where $\boldsymbol{W}_Q$, $\boldsymbol{W}_K$ and $\boldsymbol{W}_V$ are parameter weights;
$\alpha_i$ denotes the salient score for path $i$ to event pair $\boldsymbol{F}_E^{(e_1, e_2)}$;
$\boldsymbol{F}_P^{(e_1, e_2)}$ is the integrated representation of multiple paths.

\subsection{Context Encoder}

Besides the above semantic structure information, the contextual semantic information is proved to be useful for ECI~\citep{Cao2021KnowledgeEnrichedEventCausality}. 
Thus, we adopt an extra context encoder to encode the tokens of the text and obtain the contextual semantic representation of the event pair.
Specifically, we first add two pairs of special markers <e1></e1> and <e2></e2> to indicate the boundaries of the two event mentions.
Two special tokens [CLS] and [SEP] are also added to indicate the beginning and end of the whole text, respectively.
To model the representations of the tokens in the context encoder and event aggregator separately, here we adopt another BERT model to encode the context.
Following~\citet{Liu2020KnowledgeEnhancedEvent}, we use the representations of the tokens <e1> and <e2> as the representations of the two events, i.e., $e_1$ and $e_2$.
And the representation of the token [CLS] is adopted as that of the whole text.
In order to achieve sufficient interaction between the events and their corresponding contexts, we apply a linear layer and an activation function to obtain more accurate representations of the events:
\begin{equation}
    \tilde{\boldsymbol{u}}_i = \tanh (\boldsymbol{W}_u [\boldsymbol{u}_{[CLS]}|| \boldsymbol{u}_i] + \boldsymbol{b}_u),
\end{equation}
where $||$ represents the concatenation operation.
$\boldsymbol{u}_{[CLS]}$ and $\boldsymbol{u}_i$ are the representations of the whole text and $e_i$, $i \in (1,2)$, respectively. 
$\boldsymbol{W}_u$ and $\boldsymbol{b}_u$ are the weight matrix and the bias, respectively.

We again sum up the representations of the two events as the representation of the event pair:
\begin{equation}
    \boldsymbol{F}_C^{(e_1, e_2)} = \tilde{\boldsymbol{u}}_1 + \tilde{\boldsymbol{u}}_2,
\end{equation}
where $\boldsymbol{F}_C^{(e_1, e_2)}$ is the contextual representation of the event pair and will be used for further computation.

\subsection{Model Prediction}

We concatenate the representations obtained from the above three components as the final representation of each event pair:
\begin{equation}
    \boldsymbol{F}_{(e_1,e_2)} = \boldsymbol{F}_{E}^{(e_1, e_2)} || \boldsymbol{F}_{P}^{(e_1, e_2)} || \boldsymbol{F}_{C}^{(e_1, e_2)}.
\end{equation}

Then, $\boldsymbol{F}_{(e_1,e_2)}$ is fed into the softmax layer for classification,
\begin{equation}
    \boldsymbol{p} = \mathrm{softmax}(\boldsymbol{W}_f \boldsymbol{F}_{(e_1, e_2)} + \boldsymbol{b}_f),
\end{equation}
where $\boldsymbol{p}$ is the probability indicating whether there is a causal relation between two events;
$\boldsymbol{W}_f$ and $\boldsymbol{b}_f$ are trainable parameters.

\subsection{Parameter Learning}
For the classification task, the model generally adopts the cross-entropy loss function and treats all samples equally.
However, most of the samples without causality are easily predicted and these samples will dominate the total loss. 
In order to pay more attention to samples that are difficult to predict, we adopt focal loss~\citep{Lin2017FocalLossDense} as the loss function of our model:
\begin{equation}
    J(\Theta) = - \sum_{(e_i, e_j)\in E_s} \beta (1-\boldsymbol{p})^{\gamma}log(\boldsymbol{p}),
\end{equation}
where $\Theta$ denotes the model parameters;
$(e_i, e_j)$ denotes the sample in the training set $E_s$.
Besides, to balance the importance of positive and negative samples, we add the loss weighting factor $\beta \in [0,1]$ for the class ``positive'' and $1- \beta$ for the class ``negative''.

\section{Experiments}
\subsection{Datasets and Metrics}
We evaluate the proposed SemSIn on two datasets from EventStoryLine Corpus v0.9 (ESC)~\citep{Caselli2017EventStoryLineCorpus} and one dataset from Causal-TimeBank (Causal-TB)~\citep{Mirza2014AnnotatingCausalityTempEval3}, namely, ESC, ESC$^*$ and Causal-TB. 

\textbf{ESC}~\footnote{\url{https://github.com/tommasoc80/EventStoryLine}} contains 22 topics, 258 documents, and 5334 event mentions. 
The dataset is processed following~\citet{Gao2019ModelingDocumentlevelCausal}, excluding aspectual, causative, perception, and reporting event mentions, most of which are not annotated with any causality. 
After processing, there are 7805 intra-sentence event mention pairs in the corpus, of which 1770 (22.7\%) are annotated with a causal relation.
The same as previous methods~\citep{Gao2019ModelingDocumentlevelCausal, Zuo2021LearnDALearnableKnowledgeGuided}, we use the documents in the last two topics as the development set, and report the experimental results by conducting 5-fold cross-validation on the remaining 20 topics.
The dataset used in the cross-validation evaluation is partitioned as follows:
documents are sorted according to their topic IDs, which means that the training and test sets are cross-topic.
Under this setting, the data distributions of the training and test sets are inconsistent, and the generalization ability of the model is mainly evaluated.

\textbf{ESC$^*$} is another data partition setting for the ESC dataset, which is used in ~\citet{Man2022EventCausalityIdentification}. 
In this dataset, documents are randomly shuffled based on their document names without sorting according to their topic IDs. 
Thus, the training and test sets have data on all topics. Under this setting, the data distributions of the training and test sets are more consistent, and it can better reflect the performance of the model under the same distribution of data.
In real data, some causal event pairs are mostly appeared in topic-specific documents, because the event type is related to the topic of the document.
This phenomenon inspires us to split the dataset in two different ways, i.e., cross-topic partition (ESC) and random partition (ESC*). 

\textbf{Causal-TB}~\footnote{\url{https://github.com/paramitamirza/Causal-TimeBank}} contains 183 documents and 6811 event mentions.
There are 9721 intra-sentence event mention pairs in the corpus, of which 298 (3.1\%) are annotated with a causal relation.
Similar to~\citet{Liu2020KnowledgeEnhancedEvent}, we conduct 10-fold cross-validation for Causal-TB.

\textbf{Evaluation Metrics.}
For evaluation, we adopt widely used Precision (P), Recall (R), and F1-score (F1) as evaluation metrics.

\subsection{Expeimental Setup}
\textbf{Implementation Details.}
In the experiments, we use the pre-trained AMR parser  parse\_xfm\_bart\_large v0.1.0~\footnote{\url{https://github.com/bjascob/amrlib}}.
The PLM used in this paper is BERT-base~\citep{Devlin2019BERTPretrainingDeep} and it is fine-tuned during the training process.
The representation dimension of nodes and relations is set to 768, the same as the representation dimension of tokens.
The NetwokX toolkit~\footnote{\url{https://networkx.org/}} is adopted to obtain the shortest path between two events. 
The learning rate of the model is set to 1e-5 and the dropout rate is set to 0.5.
We perform grid search on the number of the RGCN layers, and it is experimentally set to 3. $\gamma$ in focal loss is set to 2. $\beta$ is set to 0.5 and 0.75 for ESC and Causal-TB, respectively.
The batch size is set to 20 for all the three datasets.
The AdamW gradient strategy is used to optimize all parameters.
Due to the sparsity of causality in the Causal-TB dataset, we use both positive and negative sampling strategies for training. 
The positive sampling rate and negative sampling rate are set to 5 and 0.3, respectively.

\begin{table}[]
    \resizebox{\linewidth}{!}{
    \begin{tabular}{l | c c c}
    \hline
    \textbf{Method} & \textbf{P} & \textbf{R} & \textbf{F1} \\
    \hline
    LSTM~\citep{Cheng2017ClassifyingTemporalRelations} & 34.0 & 41.5 & 37.4 \\
    Seq~\citep{Choubey2017SequentialModelClassifying} & 32.7 & 44.9 & 37.8 \\
    LR+~\citep{Gao2019ModelingDocumentlevelCausal} & 37.0 & 45.2 & 40.7 \\
    ILP~\citep{Gao2019ModelingDocumentlevelCausal} & 37.4 & 55.8 & 44.7 \\
    KnowDis~\citep{Zuo2020KnowDisKnowledgeEnhanced} & 39.7 & 66.5 & 49.7 \\
    MM~\citep{Liu2020KnowledgeEnhancedEvent} & 41.9 & 62.5 & 50.1 \\
    CauSeRL~\citep{Zuo2021ImprovingEventCausality} & 41.9 & 69.0 & 52.1 \\
    LSIN~\citep{Cao2021KnowledgeEnrichedEventCausality} & 47.9 & 58.1 & 52.5 \\
    LearnDA~\citep{Zuo2021LearnDALearnableKnowledgeGuided} & 42.2 & \textbf{69.8} & 52.6 \\
    \textbf{SemSIn} & \textbf{50.5} & 63.0 & \textbf{56.1} \\
    \hline
    T5 Classify$^{*}$~\citep{Man2022EventCausalityIdentification} & 39.1 & \textbf{69.5} & 47.7 \\
    GenECI$^{*}$~\citep{Man2022EventCausalityIdentification} & 59.5 & 57.1 & 58.8 \\    
    \textbf{SemSIn$^{*}$} & \textbf{64.2} & 65.7 & \textbf{64.9} \\
    \hline
    
    \end{tabular}
    }
    \caption{Experimental results on ESC and ESC$^*$.
    $^*$ denotes experimental results on ESC$^*$.
    }
    \label{table:esc-topic}
\end{table}

\begin{table}[]
    \centering
    \resizebox{\linewidth}{!}{
    \begin{tabular}{l | c c c}
    \hline
    \textbf{Method} & \textbf{P} & \textbf{R} & \textbf{F1} \\
    \hline
    RB~\citep{Mirza2014AnalysisCausalityEvents} & 36.8 & 12.3 & 18.4 \\
    DD~\citep{Mirza2014AnalysisCausalityEvents} & 67.3 & 22.6 & 33.9 \\
    VR-C~\citep{Mirza2014ExtractingTemporalCausal} & \textbf{69.0} & 31.5 & 43.2 \\
    MM~\citep{Liu2020KnowledgeEnhancedEvent} & 36.6 & 55.6 & 44.1 \\
    KnowDis~\citep{Zuo2020KnowDisKnowledgeEnhanced} & 42.3 & 60.5 & 49.8 \\
    LearnDA~\citep{Zuo2021LearnDALearnableKnowledgeGuided} & 41.9 & 68.0 & 51.9 \\
    LSIN~\citep{Cao2021KnowledgeEnrichedEventCausality} & 51.5 & 56.2 & 52.9\\
    CauSeRL~\citep{Zuo2021ImprovingEventCausality} & 43.6 & \textbf{68.1} & 53.2 \\
    GenECI~\citep{Man2022EventCausalityIdentification} & 60.1 & 53.3 & 56.5 \\
    \hline
    \textbf{SemSIn} & 52.3 & 65.8 & \textbf{58.3} \\
    \hline
    
    \end{tabular}
    }
    \caption{Experimental results on Causal-TB.}
    \label{table:ctb}
\end{table}

\textbf{Baseline Methods.}
We compare the proposed SemSIn method with two types of existing state-of-the-art (SOTA) methods, namely, feature-based ones and PLM-based ones.
For the ESC dataset, the following baselines are adopted: 
\textbf{LSTM}~\citep{Cheng2017ClassifyingTemporalRelations} is a sequential model based on dependency paths;
\textbf{Seq}~\citep{Choubey2017SequentialModelClassifying} is a sequential model that explores context word sequences;
\textbf{LR+} and \textbf{ILP}~\citep{Gao2019ModelingDocumentlevelCausal}, they consider the document-level causal structure.
For Causal-TB, the following baselines are selected: 
\textbf{RB}~\citep{Mirza2014AnalysisCausalityEvents} is a rule-based method;
\textbf{DD}~\citep{Mirza2014AnalysisCausalityEvents} is a data-driven machine learning based method;
\textbf{VR-C}~\citep{Mirza2014ExtractingTemporalCausal} is a verb rule-based model with lexical information and causal signals.

In addition, we also compare SemSIn with typical methods based on PLMs.
\textbf{KnowDis}~\citep{Zuo2020KnowDisKnowledgeEnhanced} is a knowledge enhanced distant data augmentation framework;
\textbf{MM}~\citep{Liu2020KnowledgeEnhancedEvent} is a knowledge enhanced method with mention masking generalization;
\textbf{CauSeRL}~\citep{Zuo2021ImprovingEventCausality} is a self-supervised method;
\textbf{LSIN}~\citep{Cao2021KnowledgeEnrichedEventCausality} is a method that constructs a descriptive graph to explore external knowledge;
\textbf{LearnDA}~\citep{Zuo2021LearnDALearnableKnowledgeGuided} is a learnable knowledge-guided data augmentation framework;
\textbf{T5 Classify} and \textbf{GenECI}~\citep{Man2022EventCausalityIdentification} are the methods that formulate ECI as a generation problem.

\subsection{Experimental Results}

Tables~\ref{table:esc-topic} and \ref{table:ctb} present the experimental results on the ESC and Causal-TB datasets, respectively.
Overall, our method outperforms all baselines in terms of the F1-score on both datasets. 
Compared with the SOTA methods, SemSIn achieves more than 3.5\% and 1.8\% improvement on the ESC and Causal-TB datasets, respectively.
Note that, although our method does not utilize external knowledge, it still achieves better results than the SOTA methods.
The reason is that our method makes better use of the semantic structure information in the texts.
The results indicate that the texts still contain a considerable amount of useful information for the ECI task that can be mined and exploited.

Compared with the SOTA method LearnDA in Table~\ref{table:esc-topic}, SemSIn achieves a significant improvement of 8.3\% in precision on the ESC dataset.
This suggests that SemSIn can better model the implicit associations between two events.
It can be observed that LearnDA has a higher recall score than SemSIn.
The possible reason is that LearnDA can generate event pairs out of the training set. Extra training samples make the model recall more samples and get a higher recall score.  

To verify the effectiveness of the model on the ESC$^*$ dataset, we compare the proposed method with the SOTA T5 Classify and GenECI methods. The results are in the bottom of  Table~\ref{table:esc-topic}.
SemSIn achieves 4.7\%, 8.6\%, and 6.1\% improvements of the precision, recall and F1-score, respectively. 
This again justifies that using semantic structures is beneficial for ECI. 

Comparing the results of SemSIn and SemSIn$^*$ in Table~\ref{table:esc-topic}, the experimental results under different settings have a large gap.
The results on ESC are significantly higher than those on ESC$^*$.
This is because the training and test data for ESC are cross-topic, and data on different topics usually involve diverse events.
Dealing with unseen event pairs is difficult, thus it is more challenging to conduct the ECI task on ESC than ESC$^*$.

\begin{table}[]
    \centering
    \begin{tabular}{l | c c c c}
    \hline
    \textbf{Method} & \textbf{P} & \textbf{R} & \textbf{F1} & \textbf{$\Delta$} \\
    \hline
    SemSIn$_{w/o.stru}$ & 49.8 & 49.0 & 49.4 & - \\
    SemSIn$_{w/o.path}$ & 49.3 & 52.6 & 50.9 & +1.5 \\
    SemSIn$_{w/o.cent}$ & 44.5 & 63.6 & 52.4 & +3.0 \\
    \hline
    \textbf{SemSIn} & 50.5 & 63.0 & \textbf{56.1} & \textbf{+6.7}\\
    \hline
    
    \end{tabular}
    \caption{Ablation results on ESC.
    $\Delta$ means the improvement of the F1 score relative to SemSIn$_{w/o.stru}$.
    }
    \label{table:ablation}
\end{table}

\subsection{Ablation Studies}
To illustrate the effect of two kinds of semantic structures, we conduct ablation experiments on the ESC dataset.
The results are presented in Table~\ref{table:ablation}.
${w/o.stru}$ indicates the model predicts event causality without two kinds of semantic structures.
${w/o.path}$ and ${w/o.cent}$ indicate without the event-associated structure and without the event-centric structure, respectively.

\paragraph{Impact of the Event-centric Structure.}
Compared with SemSIn, SemSIn$_{w/o.cent}$ has a 6.0\% decrease of the precision score.
By considering the event-centric structure information, the model can describe events more accurately.
Thus SemSIn$_{w/o.cent}$ is worse than SemSIn.
Comparing SemSIn$_{w/o.path}$ with SemSIn$_{w/o.stru}$, SemSIn$_{w/o.path}$ achieves 1.5\% improvements of the F1 score.
It proves that the associations between events and their arguments are vital for the ECI task.
The event-centric Structure information can enhance BERT with the ability to capture these associations.

\paragraph{Impact of the Event-associated Structure.}
Compared with SemSIn, SemSIn$_{w/o.path}$ has a 10.4\% decrease of the recall score.
This indicates that the event-associated structure information can help the model discover more causal clues between two events.
Comparing SemSIn$_{w/o.cent}$ with SemSIn$_{w/o.stru}$, SemSIn$_{w/o.cent}$ achieves 3.0\% improvements of the F1 score.
It proves that the associations between events are vital for this task.

\begin{table}[]
    \centering
    \begin{tabular}{l | c c c}
    \hline
    \textbf{Method} & \textbf{P} & \textbf{R} & \textbf{F1} \\
    \hline
    CompGCN & 46.8 & 62.0 & 53.3\\
    GCN & 50.3 & 56.8 & 53.4 \\
    RGCN & \textbf{50.5} & \textbf{63.0} & \textbf{56.1} \\
    \hline
    
    \end{tabular}
    \caption{Experimental results using different graph encoders on ESC.
    }
    \label{table:graph-encoder}
\end{table}

\subsection{Sub-module Analysis}
\paragraph{Impact of Relations in the Path.}
In the phase of acquiring semantic paths between two events, we keep only the nodes in the paths and neglect the edges.
This method achieves an F1 score of 53.3\% on ESC, which is a 2.8\% reduction compared to the model that considers both nodes and edges. 
It suggests that the relations between elements are also useful for identifying causality.

\paragraph{Impact of the Path Attention.}
In the multi-path information integration phase, we replace the attention mechanism with a method that averages the representations of multiple paths. 
This approach obtains an F1 score of 54.0\% on ESC, which is a 2.1\% reduction compared to the model utilizing the attention mechanism. 
This shows that the “Path Attention” sub-module can effectively aggregate information from multiple paths.

\subsection{Graph Encoder Analysis}
\paragraph{Impact of the Graph Encoder.}
To analyze the effect of graph encoders on experimental results, we utilized three different graph encoders, namely, GCN~\citep{Kipf2017SemiSupervisedClassificationGraph}, CompGCN~\citep{Vashishth2020CompositionbasedMultiRelationalGraph}, and RGCN~\citep{Schlichtkrull2018ModelingRelationalData}.
The results are shown in Table~\ref{table:graph-encoder}. From the results, we can observe that the best result is achieved with the model using the RGCN graph encoder.
This suggests that RGCN has the capability to utilize the edge-type information in the semantic graph more effectively, enabling more accurate aggregation of information from surrounding nodes.

\begin{figure}[t]
  \centering
  \includegraphics[width=0.9\linewidth]{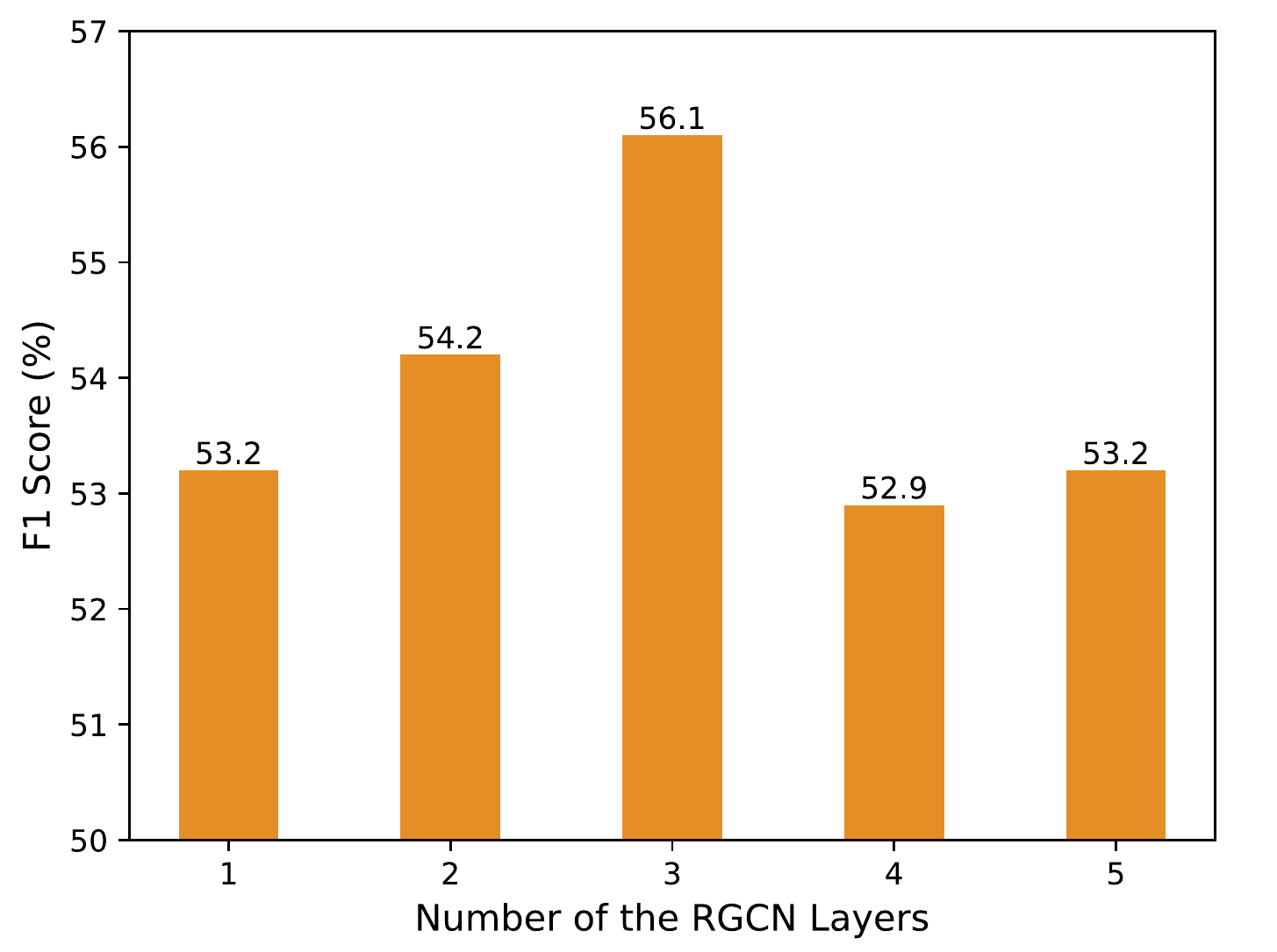}
  \caption{Impact of the number of the RGCN layers on ESC.}
  \label{fig:layers}
\end{figure}

\paragraph{Impact of the Number of the RGCN Layers.}
The number of the RGCN layers $L$ is an important parameter of the model, which means that nodes can aggregate information from their $L$-hop neighbors through message passing.
We evaluate performance of the model with different numbers of the RGCN layers on ESC.
The results are shown in Figure~\ref{fig:layers}.
A larger $L$ can get better results when $L<=3$.
This is because that events usually have associations with their arguments mentioned in the text and event arguments also have associations with some related semantic elements.
Thus introducing a relative large $L$ can describe the events more precisely.
It can be observed that the model performance decreases significantly when $L>3$. 
The reason may be that the larger $L$ may introduce some noisy nodes or the RGCN encounters the over-smoothing problem~\citep{kipf2016semi}.

\begin{figure*}[ht]
  \centering
  \includegraphics[width=\textwidth]{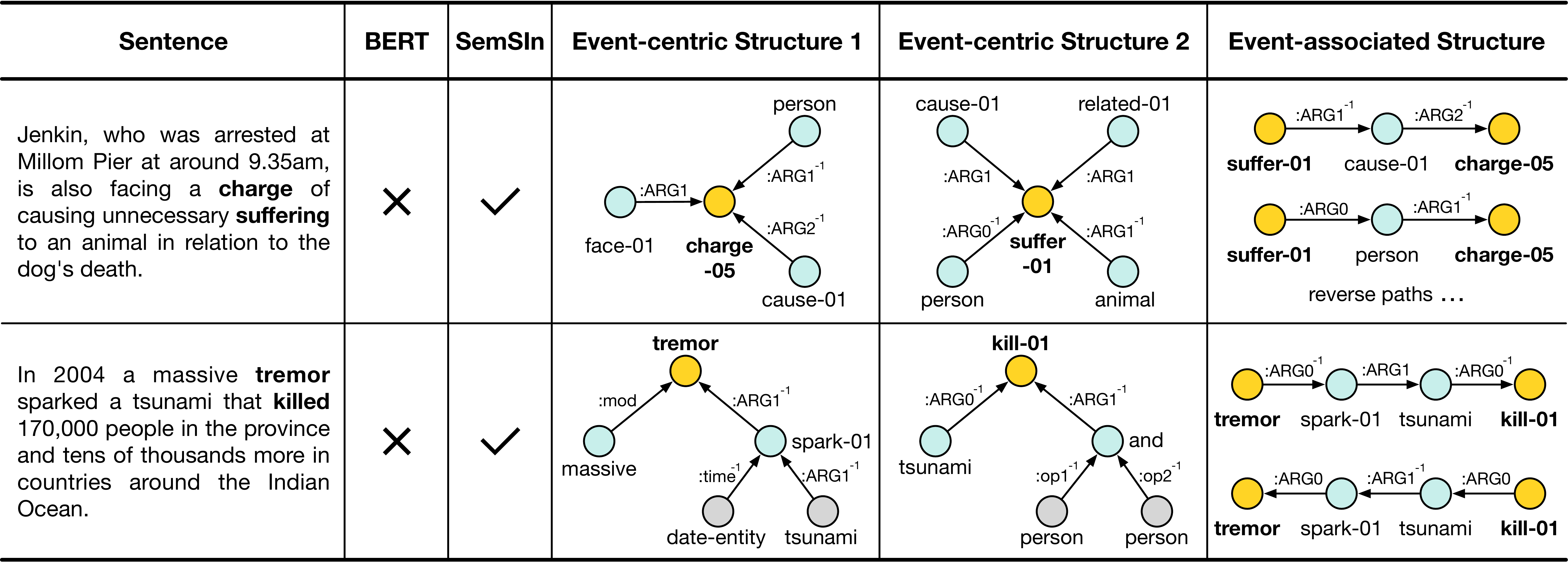}
  \caption{Results of the case study where bold indicates events. $\times$ and \checkmark \ indicate wrong and correct predictions, respectively.}
  \label{fig:case-study}
\end{figure*}

\subsection{Case Studies}
To well demonstrate how semantic structures can help improve performance, a few cases are studied.
Figure~\ref{fig:case-study} shows two cases where causal relations between events are implicit.
Here, BERT predicts the wrong answers and SemSIn predicts the correct ones, which demonstrates that leveraging the semantic structure information can effectively enhance ECI.
In Case 1, the meaning of the ``charge-05'' is that ``make an allegation or criminal charge against someone''.
Its event-centric structure information includes ``someone is facing a charge'', ``a person is charged'' and ``the charge is for causing something to happen'', which are the elements directly related to the event.
By aggregating the information of these elements, the event is semantically represented more precisely.
In Case 2, the causal relation between the two events ``tremor'' and ``kill'' is expressed indirectly through the ``tsunami'' event.
Specifically, it can be deduced using ``tremor sparked a tsunami'' and ``the tsunami killed tens of thousands of people''.
The model effectively utilizes the event-associated structure information to capture the associations between events.

\section{Conclusions}
In this paper, we proposed a new semantic structure integration model (SemSIn) for ECI, which leveraged two kinds of semantic structures, i.e., event-centric structure and event-associated structure. 
An event aggregator was utilized to aggregate event-centric structure information and a path aggregator was proposed to capture event-associated structure information between two events.
Experimental results on three widely used datasets demonstrate that introducing semantic structure information helps improve the performance of the ECI task.

\section*{Limitations}
The limitations of this work can be concluded into two points: 
(1) To obtain the associations between semantic elements, SemSIn needs to transform the texts into the corresponding semantic graphs. 
Existing methods can only transform single sentences into semantic graphs, and cannot parse texts containing multiple sentences.
Therefore, this method is not suitable for identifying causal relations between events in different sentences.
(2) SemSIn only exploits the semantic structures of the texts and does not utilize external knowledge. 
External knowledge is also important for the ECI task, and simultaneously exploiting semantic structures and external knowledge is a good direction for future studies.

\section*{Acknowledgements}
The work is supported by the National Natural Science Foundation of China under grant U1911401, the National Key Research and Development Project of China, the JCJQ Project of China, Beijing Academy of Artificial Intelligence under grant BAAI2019ZD0306, and the Lenovo-CAS Joint Lab Youth Scientist Project.
We thank anonymous reviewers for their insightful comments and suggestions.

\bibliography{custom}
\bibliographystyle{acl_natbib}

% \appendix

% \section{Example Appendix}
% \label{sec:appendix}

% This is a section in the appendix.

\end{document}